\documentclass[letterpaper]{article} %
\usepackage{aaai24}  %
\usepackage{times}  %
\usepackage{helvet}  %
\usepackage{courier}  %
\usepackage[hyphens]{url}  %
\usepackage{graphicx} %
\usepackage{natbib}  %
\usepackage{caption} %
\usepackage{algorithm}
\usepackage{algorithmic}

\usepackage{amsmath,amssymb,amsfonts}
\usepackage{multirow}
\usepackage{booktabs}
\usepackage{cleveref}
\usepackage{paralist}
\usepackage{xspace}
\usepackage{subcaption}
\usepackage{enumitem}
\usepackage{dsfont}
\usepackage{nicefrac}
\usepackage{xcolor}
\usepackage{algorithm}
\usepackage{algorithmic}

\crefname{equation}{Eq.}{Eq.}
\crefname{section}{Sec.}{Sec.}
 
\newcommand{\mscript}[1]{\text{\scriptsize{#1}}}
\newcommand{\mbf}[1]{\boldsymbol{\mathbf{#1}}}

\newcommand{\ie}{\emph{i.e.,}\xspace}

\newcommand{\eg}{\emph{e.g.,}\xspace}

\newcommand{\sig}[0]{$^\ddagger$}

\DeclareMathOperator*{\argmin}{arg\,min}
\DeclareMathOperator*{\diag}{diag}

\usepackage{newfloat}
\usepackage{listings}
\DeclareCaptionStyle{ruled}{labelfont=normalfont,labelsep=colon,strut=off} %
\floatstyle{ruled}
\newfloat{listing}{tb}{lst}{}
\floatname{listing}{Listing}
\title{
    On the Affinity, Rationality, and Diversity of Hierarchical Topic Modeling
}

\author{
    Xiaobao Wu\textsuperscript{\rm 1}, \;
    Fengjun Pan\textsuperscript{\rm 1}, \;
    Thong Nguyen\textsuperscript{\rm 2}, \;
    \\
    Yichao Feng\textsuperscript{\rm 1}, \;
    Chaoqun Liu\textsuperscript{\rm 1, 3}, \;
    Cong-Duy Nguyen\textsuperscript{\rm 1}, \;
    Anh Tuan Luu\textsuperscript{\rm 1}
}
\affiliations{
    \textsuperscript{\rm 1}Nanyang Technological University, Singapore \\
    \textsuperscript{\rm 2}National University of Singapore, Singapore \\
    \textsuperscript{\rm 3}DAMO Academy, Alibaba Group, Singapore \\
    \{xiaobao002,panf0004,yichao002,chaoqun001,nguyentr003,anhtuan.luu\}@ntu.edu.sg,\;
    e0998147@u.nus.edu

}

\usepackage{bibentry}

\begin{document}

\maketitle

\begin{abstract}
    Hierarchical topic modeling aims to discover latent topics from a corpus
    and organize them into a hierarchy to understand documents with desirable semantic granularity.
    However, existing work struggles with producing topic hierarchies of low affinity, rationality, and diversity, which hampers document understanding.
    To overcome these challenges, we in this paper propose Transport Plan and Context-aware Hierarchical Topic Model (TraCo).
    Instead of early simple topic dependencies, we propose a transport plan dependency method.
    It constrains dependencies to ensure their sparsity and balance, and also regularizes topic hierarchy building with them.
    This improves affinity and diversity of hierarchies.
    We further propose a context-aware disentangled decoder.
    Rather than previously entangled decoding,
    it distributes different semantic granularity to topics at different levels by disentangled decoding.
    This facilitates the rationality of hierarchies.
    Experiments on benchmark datasets demonstrate that our method surpasses state-of-the-art baselines, effectively improving the affinity, rationality, and diversity of hierarchical topic modeling with better performance on downstream tasks.
\end{abstract}

\section{Introduction}
    Instead of traditional flat topic models,
    hierarchical topic models strive to discover a topic hierarchy from documents \cite{griffiths2003hierarchical,teh2004sharing}.
    Each topic is interpreted as relevant words to represent a semantic concept.
    The hierarchy captures the relationships among topics and organizes them by semantic granularity:
    child topics at lower levels are relatively specific to parent topics at higher levels.
    Therefore hierarchical topic models can provide a more comprehensive understanding of complex documents with desirable granularity.
    Due to this advantage,
    they have been applied in various downstream applications like
    document retrieval \cite{weninger2012document},
    sentiment analysis \cite{kim2013hierarchical},
    and text summarization \cite{celikyilmaz2010hybrid} or
    generation \cite{guo2020recurrent,tuan2020capturing}.

    \begin{figure}[!t]
    \centering
    \includegraphics[width=0.9\linewidth]{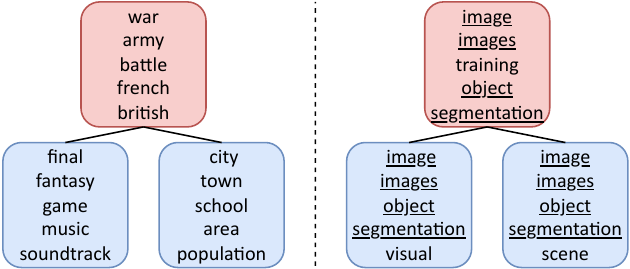}
    \caption{
        Illustration of low affinity (left), and low rationality and diversity issues (right) from Wikitext-103 and NeurIPS.
        Each rectangle is the top related words of a topic from HyperMiner \cite{xu2022hyperminer}.
        Repetitive words are underlined.
    }
    \label{fig_motivation}
\end{figure}

    Existing hierarchical topic models have two categories.
    The first category is conventional models like hLDA \cite{griffiths2003hierarchical} and its variants \cite{kim2012recursive,paisley2013nested}.
    They infer parameters through Gibbs sampling or Variational Inference.
    But they cannot well handle large-scale datasets due to their high computational cost \cite{chen2021tree,chen2023nonlinear}.
    The second category is neural models including HNTM \cite{chen2021hierarchical}, HyperMiner \cite{xu2022hyperminer}, and others \cite{isonuma2020tree,chen2021tree,chen2023nonlinear,duan2021sawtooth}.
    They generally
    follow VAE frameworks and enjoy back-propagation for faster parameter inferences \cite{wu2023survey}.

    However, these work struggles with producing low-quality topic hierarchies due to three issues:
    \begin{inparaenum}[(\bgroup\bfseries i\egroup)]
        \item
            \emph{Low Affinity}:
            child topics are \emph{not} affinitive to their parents \cite{kim2012recursive}.
            As exemplified in the left of \Cref{fig_motivation}, the parent topic relates to ``army'',
            whereas its child topics contain irrelevant words ``game music'' and ``school''.
            Such low-affinity hierarchies capture inaccurate relationships among topics.
        \item
            \emph{Low Rationality}:
            child topics are excessively similar to their parent topics instead of being specific to them as expected \cite{viegas2020cluhtm}.
            The right part of \Cref{fig_motivation}
            shows the parent and its child topics all focus on ``image segmentation'' with the same granularity.
            So low-rationality hierarchies provide topics with less comprehensive granularity.
        \item
            \emph{Low Diversity}: sibling topics are repetitive instead of being diverse as expected \cite{zhang2022nonparametric}.
            In the right part of \Cref{fig_motivation}, the two sibling topics repeat each other and become redundant,
            implying other undisclosed latent topics.
            Thus low-diversity hierarchies produce less informative and incomplete topics.
    \end{inparaenum}
    Due to these issues,
    existing hierarchical topic models generate low-quality hierarchies, which impedes document understanding and thus damages their interpretability and performance on downstream applications.

    To address these challenges,
    we in this paper propose a novel neural hierarchical topic model,
    called \textbf{Tra}nsport Plan and \textbf{Co}ntext-aware Hierarchical Topic Model (\textbf{TraCo}).
    First, to address the low affinity and diversity issues,
    we propose a new \textbf{Transport Plan Dependency} (\textbf{TPD}) approach.
    Instead of unconstrained dependencies as previous work \cite{chen2021tree,duan2021sawtooth,xu2022hyperminer},
    TPD models dependencies of hierarchical topics as optimal transport plans between them,
    which constrains the dependencies to ensure their sparsity and balance.
    Guided by the constrained dependencies,
    TPD additionally regularizes the building of topic hierarchies:
    it pushes a child topic only close to its parent and away from others, and avoids gathering excessive sibling topics together.
    As a result, this improves the affinity between child and parent topics and the diversity of sibling topics in learned hierarchies.

    Second, to solve the low rationality issue, we further propose a novel \textbf{Context-aware Disentangled Decoder} (\textbf{CDD}).
    Rather than entangled decoding in early work \cite{chen2021hierarchical,chen2023nonlinear,li2022alleviating},
    CDD decodes input documents using topics at each level individually, leading to disentangled decoding.
    In addition, the decoding of each level incorporates a bias containing topical semantics from its contextual levels.
    This incorporation forces topics at each level to cover semantics different from their contextual levels.
    In consequence, CDD can distribute different semantic granularity to topics at different levels,
    which therefore enhances the rationality of hierarchies.
    We conclude the contributions of this paper as follows~\footnote{Our code is available at \url{https://github.com/bobxwu/TraCo}.}:
    \begin{itemize}[leftmargin=*]
        \item
            We propose a novel neural hierarchical topic model with a new transport plan dependency method
            that regularizes topic hierarchy building
            with sparse and balanced dependencies,
            mitigating the low affinity and diversity issues.
        \item
            We further propose a new context-aware disentangled decoder,
            which explicitly distributes different semantic granularity to topics at different levels and thus alleviates the low rationality issue.
        \item
            We conduct extensive experiments on benchmark datasets and demonstrate that our model surpasses state-of-the-art baselines and significantly improves the affinity, rationality, and diversity of topic hierarchies.
    \end{itemize}

\section{Related Work}

    \paragraph{Conventional Hierarchical Topic Models}
        Instead of flat topics
        like LDA
        \cite{blei2003latent,Wu2019},
        \citet{griffiths2003hierarchical} propose hLDA to generate topic hierarchies with a nested Chinese Restaurant Process (nCRP).
        To relieve its single-path formulation,
        \citet{paisley2013nested} propose a nested Hierarchical Dirichlet Process.
        More variants are explored \cite{mimno2007mixtures,blei2010nested,perotte2011hierarchically,kim2012recursive}.
        Alternatively, \citet{viegas2020cluhtm} use NMF \cite{liu2018topic} with cluster word embeddings;
        \citet{shahid2023hyhtm} extend it by hyperbolic word embeddings.
        But they cannot infer topic distributions of documents.

    \paragraph{Neural Hierarchical Topic Models}
        Recently, neural hierarchical topic models have emerged in the framework of VAE \cite{Kingma2014a,Rezende2014,nguyen2021contrastive,Wu2020short,wu2022mitigating,wu2023infoctm,wu2023topmost}.
        Some follow conventional models \cite{pham2021neural,zhang2022nonparametric}.
        \citet{isonuma2020tree} first propose a tree-structure topic model with two simplified doubly-recurrent neural networks.
        \citet{chen2021tree} propose nTSNTM with a stick-breaking process prior.
        Lately parametric settings attract more attention, \ie specify the number of topics at each level of a hierarchy \cite{wang2022knowledge,wang2023hierarchical}.
        \citet{chen2021hierarchical} propose a manifold regularization on topic dependencies.
        \citet{li2022alleviating} use skip-connections for
        decoding and train with a policy gradient approach.
        \citet{xu2022hyperminer} model topic and word embeddings in hyperbolic space.
        \citet{chen2023nonlinear} use a Gaussian mixture prior and nonlinear structural equations to model dependencies.
        We follow the popular parametric setting,
        but differently focus on the low affinity, rationality, and diversity issues of hierarchical topic modeling.
        To address these issues,
        we propose the transport plan dependency to regularize topic hierarchy building
        and the context-aware disentangled decoder to separate semantic granularity.

\section{Methodology}
    In this section, we recall the problem setting and notations of hierarchical topic modeling.
    Then we propose our transport plan dependency method and context-aware disentangled decoder.
    Finally we present our \textbf{Tra}nsport Plan and \textbf{Co}ntext-aware Hierarchical Topic Model (\textbf{TraCo}).

    \subsection{Problem Setting and Notations}
        Consider a collection of $N$ documents: $\{\mbf{x}^{(1)}, \dots, \mbf{x}^{(N)}\}$
        with $V$ unique words (vocabulary size).
        Following \citet{chen2021hierarchical,duan2021sawtooth},
        we aim to discover a topic hierarchy with $L$ levels from this collection, where level $\ell$ has $K^{(\ell)}$ latent topics.
        We build this hierarchy with dependency matrices
        describing the hierarchical relations between topics at two levels.
        For example,
        $\mbf{\varphi}^{(\ell)} \!\! \in \!\! \mathbb{R}^{K^{(\ell+1)} \! \times \! K^{(\ell)}} $ denotes the dependency matrix between topics at level $\ell$ and $\ell \!+\! 1$,
        where $\varphi^{(\ell)}_{kk'}$ is the relation between Topic\#$k$ at level $\ell \!+\! 1$ and Topic\#$k'$ at level $\ell$.
        Child topics should have high dependencies on their parents and low on others.
        Following LDA,
        we define each latent topic as a distribution over words (topic-word distribution),
        \eg Topic\#$k$ at level $\ell$ is defined as $\mbf{\beta}_{k}^{(\ell)} \!\! \in \!\! \mathbb{R}^{V} $.
        Then $\mbf{\beta}^{(\ell)} \!\!=\!\! ( \mbf{\beta}^{(\ell)}_{1},\dots,\mbf{\beta}^{(\ell)}_{K^{(\ell)}} ) \!\! \in \!\! \mathbb{R}^{ V \!\times\! K^{(\ell)} } $ is the topic-word distribution matrix of level $\ell$.
        In addition, we infer doc-topic distributions at each level,
        \ie topic proportions in a document.
        For example, we denote $\mbf{\theta}^{(\ell)} \!\! \in \!\! \Delta_{K^{(\ell)}}$ as the doc-topic distribution of a document $\mbf{x}$ at level $\ell$, where $\Delta_{K^{(\ell)}}$ is a probability simplex.

\begin{figure}[!t]
    \centering
    \begin{subfigure}[b]{0.33333\linewidth}
        \centering
        \includegraphics[width=\linewidth]{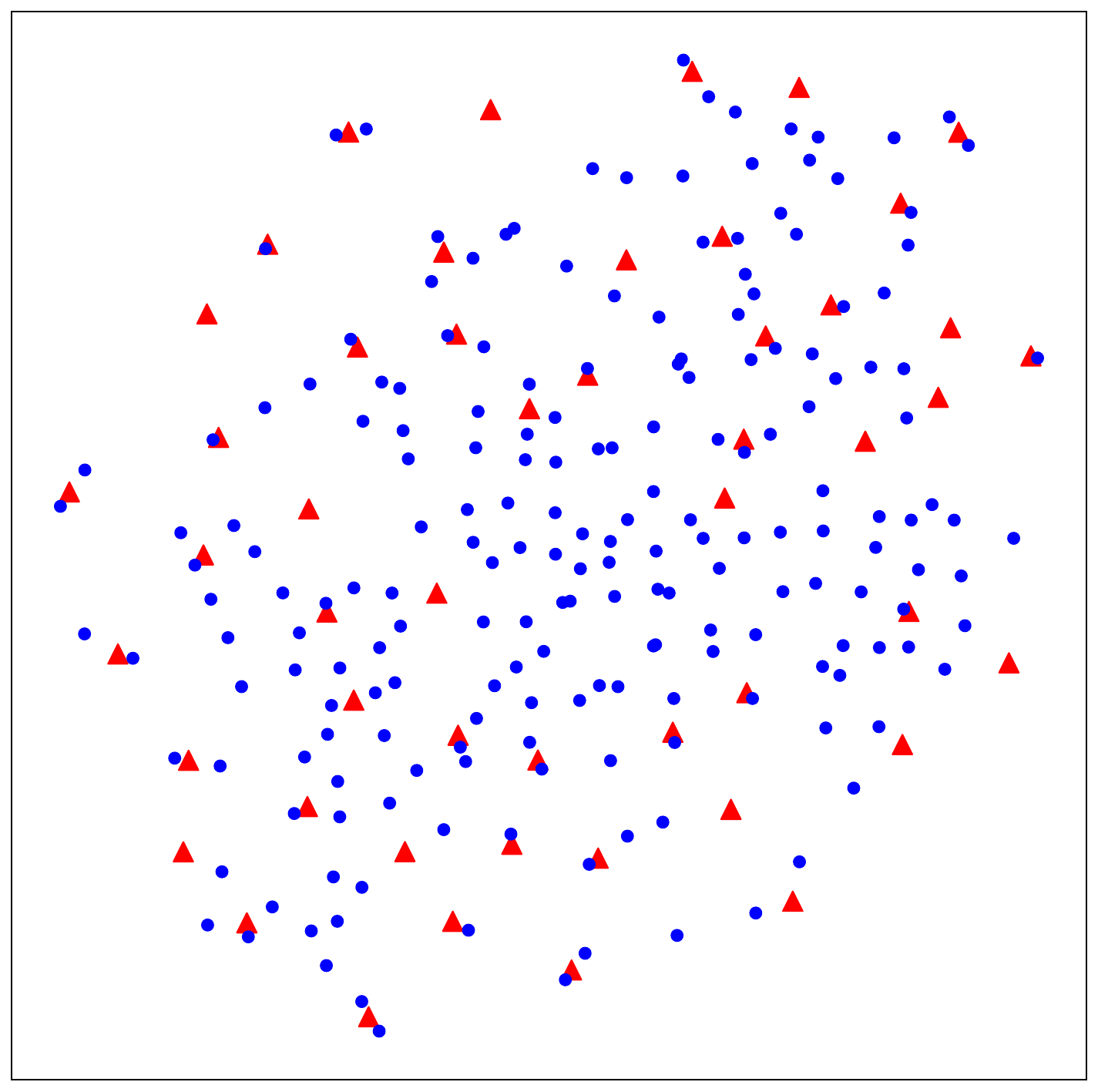}
        \caption{HyperMiner}
        \label{fig_visualization_HyperMiner}
    \end{subfigure}%
    \begin{subfigure}[b]{0.33333\linewidth}
        \centering
        \includegraphics[width=\linewidth]{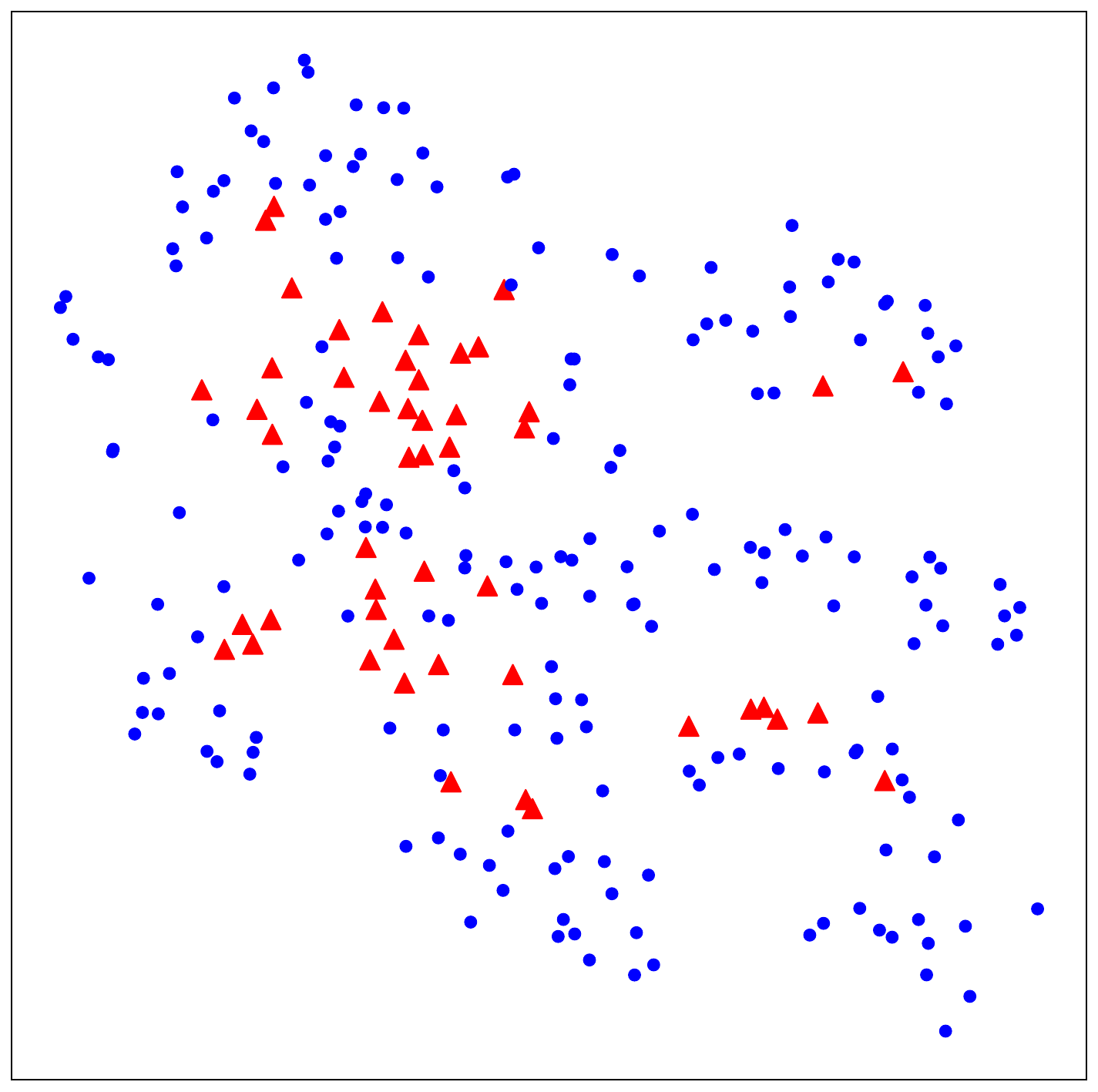}
        \caption{NGHTM}
        \label{fig_visualization_NGHTM}
    \end{subfigure}%
    \begin{subfigure}[b]{0.33333\linewidth}
        \centering
        \includegraphics[width=\linewidth]{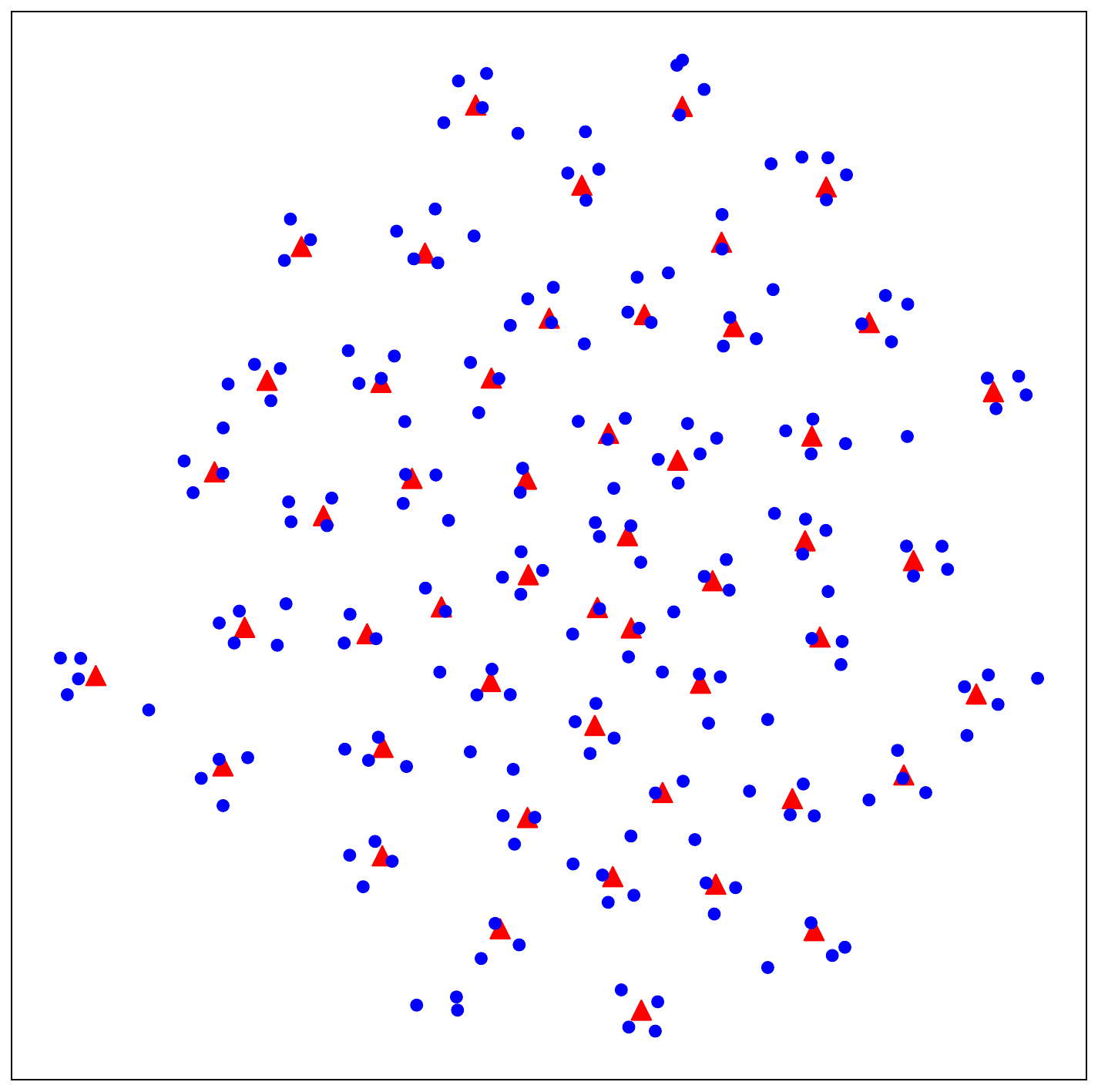}
        \caption{TraCo}
        \label{fig_visualization_TraCo}
    \end{subfigure}%
    \caption{
        t-SNE visualization \cite{Maaten2008}
        of learned child ({\color{blue}$\bullet$}) and parent ({\color{red}$\blacktriangle$}) topic embeddings of two levels.
        (\textbf{a,b}): Some child topic embeddings are \emph{not} close enough to their parents; some are excessively gathered together.
        (\textbf{c}): TraCo pushes each child topic embedding only close to its parent and away from others, and avoids gathering excessive ones together.
    }
    \label{fig_visualization}
\end{figure}

    \subsection{Parameterizing Hierarchical Latent Topics}
        At first we parameterize hierarchical latent topics.
        Following \citet{Miao2017,dieng2020topic},
        we project both words in the vocabulary and topics at all levels into an embedding space.
        In detail,
        we have $V$ word embeddings: $\mbf{W} \!\!=\!\! (\mbf{w}_{1}, \dots, \mbf{w}_{V}) \! \in \! \mathbb{R}^{D \times V}$
        where $D$ is the dimension.
        Similarly,
        we have $K^{(\ell)}$ topic embeddings for level $\ell$:
        $ \mbf{T}^{(\ell)} \!\!=\!\! ( \mbf{t}_{1}^{(\ell)}, \dots, \mbf{t}_{K^{(\ell)}}^{(\ell)} ) \! \in \! \mathbb{R}^{ D \times K^{(\ell)} } $.
        Each topic (word) embedding represents its semantics.
        To model latent topics at level $\ell$,
        we calculate its topic-word distribution matrix $\mbf{\beta}^{(\ell)}$ following \citet{wu2023effective} as
        \begin{align}
            \beta^{(\ell)}_{k,i} = \frac{ \exp ( -\| \mbf{t}^{(\ell)}_{k} - \mbf{w}_{i} \|^2 / \tau) }{ \sum_{k'=1}^{K} \exp ( -\| \mbf{t}^{(\ell)}_{k'} - \mbf{w}_{i} \|^2 / \tau) } \label{eq_beta}
        \end{align}
        where ${\beta}^{(\ell)}_{k,i}$ is the correlation between $i$-th word and Topic\#$k$ at level $\ell$
        with $\tau$ as a
        hyperparameter.
        Here we model the correlation as the Euclidean distance between word and topic embeddings
        and normalize over all topics at level $\ell$.

    \subsection{Transport Plan Dependency}
        In this section
        we analyze why topic hierarchies are of low affinity and diversity,
        and then propose a novel solution called the Transport Plan Dependency (TPD).

        \paragraph{Why Low Affinity and Diversity?}
            As illustrated in \Cref{fig_motivation},
            previous models struggle with the low affinity and diversity issues.
            We consider the reason lies in their ways of modeling topic dependencies.
            Specifically, previous methods model dependencies between topics as the similarities between their topic embeddings.
            For instance, most studies compute the dot-product of topic embeddings as similarities and normalize them with a softmax function \cite{chen2021tree,duan2021sawtooth}.
            However, these dependencies are unconstrained and cannot regularize the building of topic hierarchies.
            As shown in \Cref{fig_visualization_HyperMiner,fig_visualization_NGHTM},
            this incurs the low affinity and diversity issues:
            \begin{inparaenum}[(\bgroup\bfseries i\egroup)]
                \item
                    \textbf{The dependencies may lack sparsity},
                    indicating child topic embeddings are \emph{not} close enough to their parents.
                    As a result, child topics are insufficiently associated with their parent topics,
                    which damages the affinity of hierarchies.
                \item
                    \textbf{The dependencies could be imbalanced},
                    indicating excessive child topic embeddings are gathered together close to only a few parents.
                    In consequence, these topics become siblings and contain similar semantics,
                    which impairs the diversity of hierarchies.
            \end{inparaenum}

\begin{figure}[!t]
    \centering
    \includegraphics[width=0.95\linewidth]{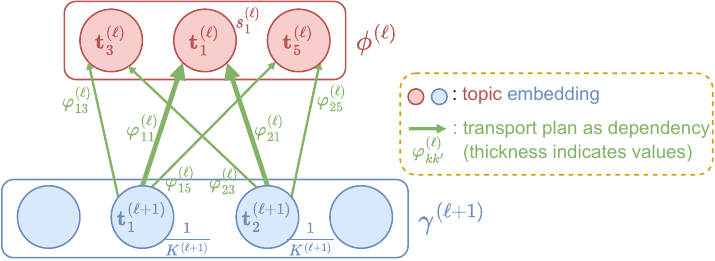}
    \caption{
        Illustration of
        TPD.
        It models the dependency $\mbf{\varphi}^{(\ell)}_{kk'}$~as the transport plan
        from topic embedding $\mbf{t}^{(\ell+1)}_{k}$ to $\mbf{t}^{(\ell)}_{k'}$
        in measures $\gamma^{(\ell+1)}$ and $\phi^{(\ell)}$,
        constrained by the weight of $\mbf{t}^{(\ell+1)}_{k}$ as $\nicefrac{1}{K^{(\ell+1)}}$ and $\mbf{t}^{(\ell)}_{k'}$ as $s^{(\ell)}_{k'}$.
        Here TPD pushes $\mbf{t}^{(\ell+1)}_1$ close to $\mbf{t}^{(\ell)}_{1}$ and away from others, similar for $\mbf{t}^{(\ell+1)}_{2}$.
    }
    \label{fig_illustration_TPD}
\end{figure}

\begin{figure*}[!t]
    \centering
    \begin{subfigure}[b]{0.29824\linewidth}
        \centering
        \includegraphics[width=0.7\linewidth]{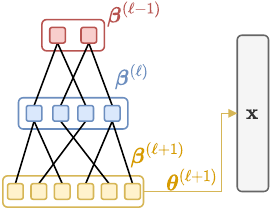}
        \caption{Lowest-Level Decoder}
        \label{fig_decoder_lowest-level}
    \end{subfigure}%
    \begin{subfigure}[b]{0.29824\linewidth}
        \centering
        \includegraphics[width=0.7\linewidth]{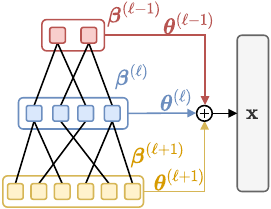}
        \caption{Aggregation Decoder}
        \label{fig_decoder_aggregation}
    \end{subfigure}%
    \begin{subfigure}[b]{0.40351\linewidth}
        \centering
        \includegraphics[width=0.7\linewidth]{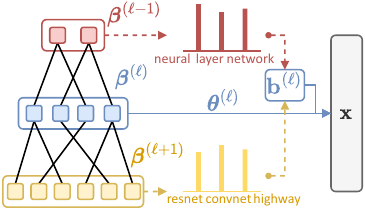}
        \caption{Context-aware Disentangled Decoder}
        \label{fig_decoder_disentangled}
    \end{subfigure}%
    \caption{
        Comparison of decoders for hierarchical topic modeling.
        Here $\mbf{\beta}^{(\ell)}$ and $\mbf{\theta}^{(\ell)}$ are the topic-word distribution matrix and doc-topic distribution at level $\ell$ respectively.
        $\mbf{x}$ is an input document to be decoded.
        (a):
            Decoding only with the lowest level.
        (b):
            Decoding with all levels.
        (c):
            Decoding with each level individually.
            For example,
            here the decoding using level $\ell$ incorporates the contextual topical bias $\mbf{b}^{(\ell)}$.
            The bias includes topical semantics from contextual levels ($\ell\!-\!1$ and $\ell\!+\!1$), like the top related words ``neural layer network'' and ``resnet convnet highway''.
            This encourages topics at level $\ell$ ($\mbf{\beta}^{(\ell)}$) to cover semantics different from them, like ``deep convolutional cnn''
            (See this example in case studies).
            It is similar for other levels.
    }
    \label{fig_decoder}
\end{figure*}

        \paragraph{Modeling Dependencies as Transport Plans}
            Based on the above analysis,
            to solve the low affinity and diversity issues,
            we propose a new Transport Plan Dependency (TPD) method that regularizes topic hierarchy building with sparse and balanced dependencies.
            \Cref{fig_illustration_TPD} illustrates TPD, and \Cref{fig_visualization_TraCo} shows its effectiveness.

            To constrain dependencies,
            we model them as the transport plan of a particularly defined optimal transport problem.
            Specifically,
            we define discrete measures on the topic embeddings at levels $\ell \!+\! 1$ and $\ell$ respectively as
            $
                \gamma^{(\ell+1)} \!\!=\!\! \sum_{k=1}^{K^{(\ell+1)}} \!\!\! \nicefrac{1}{K^{(\ell+1)}} \sigma_{\mbf{t}^{(\ell+1)}_{k}}
            $
            and
            $
                \phi^{(\ell)} \!\!=\!\! \sum_{k'=1}^{K^{(\ell)}} s^{(\ell)}_{k'} \sigma_{\mbf{t}^{(\ell)}_{k'}}
            $,
            where $\sigma_{x}$ denotes the Dirac unit mass on $x$.
            Here the measures specify the weight of each topic embedding at level $\ell \!+\! 1$ as $\nicefrac{1}{K^{(\ell+1)}}$,
            and each at level $\ell$ as $s^{(\ell)}_{k'}$
            where
            $\mbf{s}^{(\ell)} \!\!=\!\! (s^{(\ell)}_{1},\dots, s^{(\ell)}_{K^{(\ell)}})$ is a weight vector
            and its sum is $1$.
            Then we formulate an entropic regularized optimal transport problem between them:
            \begin{align}
                & \argmin_{\mbf{\pi}^{(\ell)} \in \mathbb{R}_{+}^{K^{(\ell+1)} \times K^{(\ell)}}} \!\!\!\!\! \mathcal{L}_{\mscript{OT}_{\varepsilon}}( \gamma^{(\ell+1)} \!\!, \phi^{(\ell)} ), \quad \text{where} \notag \\
                & \mathcal{L}_{\mscript{OT}_{\varepsilon}}( \gamma^{(\ell+1)} \!\!, \phi^{(\ell)} ) \!\!=\!\!\!\!\!  \sum_{k=1}^{K^{(\ell+1)}} \!\! \sum_{k'=1}^{K^{(\ell)}} \!\! C^{(\ell)}_{kk'} \pi^{(\ell)}_{kk'} + \varepsilon \pi^{(\ell)}_{kk'} ( \log \pi^{(\ell)}_{kk'} \!\!-\!\! 1) \notag \\
                & \mathrm{s.t.} \; \mbf{\pi}^{(\ell)} \mathds{1}_{K^{(\ell)}} \!\!=\!\! \nicefrac{1}{K^{(\ell+1)}} \mathds{1}_{K^{(\ell+1)}}
                ,
                (\mbf{\pi}^{(\ell)})^{\top} \mathds{1}_{K^{(\ell+1)}} \!\!=\!\! \mbf{s}^{(\ell)} . \label{eq_OT_hierarchical_topic}
            \end{align}
            The first term of $\mathcal{L}_{\mscript{OT}_{\varepsilon}}$ is the original optimal transport problem,
            and the second term is the entropic regularization with hyperparameter $\varepsilon$ to make this problem tractable \cite{canas2012learning}.
            \Cref{eq_OT_hierarchical_topic} is to find a transport plan $\mbf{\pi}^{(\ell)}$ that minimizes the total cost of transporting the weights of topic embeddings at level $\ell\!+\!1$ to topic embeddings at $\ell$ and fulfills the two constraints.
            Here $\pi^{(\ell)}_{kk'}$ indicates the transport weight from $\mbf{t}^{(\ell+1)}_{k}$ to $\mbf{t}^{(\ell)}_{k'}$,
            and we compute the transport cost between them 
            as Euclidean distance: $C^{(\ell)}_{kk'} \!\!=\!\! \| \mbf{t}^{(\ell+1)}_{k} \!-\! \mbf{t}^{(\ell)}_{k'} \|^{2}$.
            We denote $\mbf{C}^{(\ell)}$ as the transport cost matrix.
            \Cref{eq_OT_hierarchical_topic} has two constraints on $\mbf{\pi}^{(\ell)}$ to balance transport weights where $\mathds{1}_{K}$ is a $K$-dimensional column vector of ones.

            To ensure the sparsity and balance of dependencies,
            we model them as the optimal transport plan solution of \Cref{eq_OT_hierarchical_topic}:
            \begin{align}
                \mbf{\varphi}^{(\ell)}
                = \mathrm{sinkhorn} ( \mathcal{L}_{\mscript{OT}_{\varepsilon}} ( \gamma^{(\ell+1)} \!\!, \phi^{(\ell)} ) )
                .
            \end{align}
            We resort to Sinkhorn's algorithm \cite{sinkhorn1964relationship,cuturi2013lightspeed} to approximate the optimal transport plan 
            (See details in Appendix A).
            This makes the obtained $\mbf{\varphi}^{(\ell)}$ a differentiable variable parameterized by transport cost matrix $\mbf{C}$ \cite{salimans2018improving,genevay2018learning}.
            Here to obtain sparse and balanced dependencies,
            we model the dependency between Topic\#$k$ at level $\ell \!+\! 1$ and Topic\#$k'$ at level $\ell$
            as the transport weight between their topic embeddings $\mbf{t}^{(\ell+1)}_{k}$ and $\mbf{t}^{(\ell)}_{k'}$.
            Early studies prove that the optimal transport plan becomes sparse under a small $\varepsilon$ \cite{peyre2019computational,Genevay2019Different}.
            Therefore the modeled dependencies can keep sparsity.
            Besides,
            the two constraints in \Cref{eq_OT_hierarchical_topic} ensure that
            the sparse transport plan needs to transport multiple topic embeddings at level $\ell \!+\! 1$ with a total weight of $s^{(\ell)}_{k'}$
            to topic embedding $\mbf{t}^{(\ell)}_{k'}$ at level $\ell$.
            Thus the modeled dependencies under these constraints can maintain balance.

        \paragraph{Objective for TPD}
            To regularize topic hierarchy building,
            we formulate the objective for TPD with the dependencies:
            \begin{align}
                \mathcal{L}^{(\ell)}_{\mscript{TPD}} = \sum_{k=1}^{K^{(\ell+1)}} \sum_{k'=1}^{K^{(\ell)}} C_{kk'} \varphi^{(\ell)}_{kk'}
                \label{eq_TPD}
            \end{align}
            where we minimize the total distance between topic embeddings at two levels weighted by dependencies.
            As shown in \Cref{fig_illustration_TPD},
            since dependencies $\mbf{\varphi}^{(\ell)}$ are sparse,
            \Cref{eq_TPD} pushes a child topic embedding only close to its parent and away from others.
            This facilitates the affinity of learned hierarchies.
            As the dependencies are also balanced, it properly aggregates child topic embeddings and avoids gathering excessive ones together.
            This improves the diversity of learned hierarchies.
            We demonstrate these in ablation studies.

    \subsection{Inferring Doc-Topic Distributions of Levels} \label{sec_doc-topic_distributions}
        We infer doc-topic distributions over each level for document decoding.
        We first infer $\mbf{\theta}^{(L)}$, the doc-topic distributions over topics at the lowest level $L$ following normal topic models \cite{Srivastava2017,Wu2020,Wu2021discovering}.
        In detail,
        we define a random variable $\mbf{r} \!\! \in \!\! \mathbb{R}^{K^{(L)}}$ with a logistic normal prior $\mathcal{LN}(\mbf{\mu}_0, \mbf{\Sigma}_0)$
        where $\mbf{\mu}_0$ and $\mbf{\Sigma}_0$ are the mean and diagonal covariance matrix.
        We model its variational distribution as $q_{\Theta}(\mbf{r}|\mbf{x}) = \mathcal{N}(\mbf{\mu}, \mbf{\Sigma})$.
        To model parameters $\mbf{\mu}, \mbf{\Sigma}$,
        we use a neural network encoder $f_{\Theta}$ parameterized by $\Theta$ with the Bag-of-Words of document $\mbf{x}$ as inputs.
        Then we sample $\mbf{r}$ via the reparameterization trick as
        $
            \mbf{r} \!\!=\!\! \mbf{\mu} + (\mbf{\Sigma})^{1/2} \mbf{\epsilon}
        $
        where $\mbf{\epsilon} \sim \mathcal{N}(\mbf{0}, \mbf{I})$.
        We compute $\mbf{\theta}^{(L)}$ with a softmax function as
        $
            \mbf{\theta}^{(L)} \!\!=\!\! \mathrm{softmax}(\mbf{r})
        $.
        Thereafter, we infer doc-topic distributions of a higher level $\ell$ as
        \begin{align}
            \mbf{\theta}^{(\ell)} = \bigl( \prod_{\ell'=\ell}^{L-1} (K^{(\ell'+1)} \mbf{\varphi}^{(\ell')})^{\top} \bigr) \mbf{\theta}^{(L)}
            \;\; \text{where} \;\; l < L.
            \label{eq_theta}
        \end{align}
        Here we transform $\mbf{\theta}^{(L)}$ via the dependencies of each level,
        and the multiplication of $K^{(\ell'+1)}$
        rescales $\mbf{\varphi}^{(\ell')}$ to produce normalized doc-topic distribution $\mbf{\theta}^{(\ell)}$.

\begin{table*}[ht]
    \centering
    \setlength{\tabcolsep}{2mm}
    \renewcommand{\arraystretch}{1.1}
    \resizebox{0.65\linewidth}{!}{
    \begin{tabular}{lrrrrrrrrrrrrrr}
        \toprule
        \multirow{2}[4]{*}{Model} & \multicolumn{2}{c}{NeurIPS} &       & \multicolumn{2}{c}{ACL} &       & \multicolumn{2}{c}{NYT} &       & \multicolumn{2}{c}{Wikitext-103} &       & \multicolumn{2}{c}{20NG} \\
        \cmidrule{2-3}\cmidrule{5-6}\cmidrule{8-9}\cmidrule{11-12}\cmidrule{14-15}      & \multicolumn{1}{c}{TC} & \multicolumn{1}{c}{TD} &       & \multicolumn{1}{c}{TC} & \multicolumn{1}{c}{TD} &       & \multicolumn{1}{c}{TC} & \multicolumn{1}{c}{TD} &       & \multicolumn{1}{c}{TC} & \multicolumn{1}{c}{TD} &       & \multicolumn{1}{c}{TC} & \multicolumn{1}{c}{TD} \\
        \midrule
        nTSNTM & 0.389\sig & 0.050\sig &       & 0.369\sig & 0.077\sig &       & 0.348\sig & 0.064\sig &       & 0.366\sig & 0.210\sig &       & 0.361\sig & 0.118\sig \\
        HNTM  & 0.356\sig & 0.118\sig &       & 0.368\sig & 0.191\sig &       & 0.345\sig & 0.134\sig &       & 0.383\sig & 0.300\sig &       & 0.376\sig & 0.151\sig \\
        NGHTM & 0.362\sig & 0.235\sig &       & 0.371\sig & 0.235\sig &       & 0.359\sig & 0.145\sig &       & 0.391\sig & 0.224\sig &       & 0.358\sig & 0.166\sig \\
        SawETM & 0.367\sig & 0.586\sig &       & 0.362\sig & 0.483\sig &       & 0.374\sig & 0.523\sig &       & 0.392\sig & 0.348\sig &       & 0.366\sig & 0.430\sig \\
        DCETM & 0.385\sig & 0.173\sig &       & 0.398\sig & 0.090\sig &       & 0.371\sig & 0.652\sig &       & 0.385\sig & 0.086\sig &       & 0.380\sig & 0.561\sig \\
        ProGBN & 0.377\sig & 0.436\sig &       & 0.372\sig & 0.462\sig &       & 0.372\sig & 0.502\sig &       & 0.400\sig & 0.542\sig &       & 0.370\sig & 0.495\sig \\
        HyperMiner & 0.374\sig & 0.632\sig &       & 0.364\sig & 0.636\sig &       & 0.365\sig & 0.580\sig &       & 0.390\sig & 0.443\sig &       & 0.359\sig & 0.456\sig \\
        \midrule
        \textbf{TraCo} & \textbf{0.438} & \textbf{0.824} &       & \textbf{0.421} & \textbf{0.823} &       & \textbf{0.401} & \textbf{0.782} &       & \textbf{0.407} & \textbf{0.813} &       & \textbf{0.394} & \textbf{0.718} \\
        \bottomrule
    \end{tabular}%
    }
    \caption{
        Topic quality results of Topic Coherence (TC) and Diversity (TD).
        The best are in bold.
        The superscript $\ddagger$ means the gain of is statistically significant at 0.05 level.
    }
    \label{tab_topic_quality}%
\end{table*}%

    \subsection{Context-aware Disentangled Decoder}
        In this section we explore why the low rationality issue happens.
        Then we propose a novel Context-aware Disentangled Decoder (CDD)
        to address this issue.

        \paragraph{Why Low Rationality?}
            As exemplified in \Cref{fig_motivation},
            early methods suffer from low rationality,
            \ie child topics have the same granularity as parent topics instead of being specific to them.
            We conceive the underlying reason lies in their decoders.
            As shown in \Cref{fig_decoder}, previous decoders can be classified into two types.
            The first type is \textbf{lowest-level decoders} \cite{duan2021sawtooth,xu2022hyperminer}.
            Their decoding only engages the lowest-level topics.
            Higher-level topics are the linear combinations of these lowest-level topics via dependency matrices.
            In consequence,
            this entangles topics at all levels to cover the same semantic granularity, causing low rationality.
            The second type is \textbf{aggregation decoders} \cite{chen2021tree,chen2021hierarchical,li2022alleviating,chen2023nonlinear}.
            Their decoding involves all levels, which still entangles topics at all levels.
            This endows the same semantics to these topics, so they become relevant but have similar granularity.
            As a result,
            learned hierarchies tend to have low rationality even with high affinity.
            Recently \citet{duan2023bayesian} craft documents with more related words for the decoding of higher levels,
            but their granularity cannot be separated, still experiencing low rationality.
            See supports in the experiment section.

        \paragraph{Contextual Topical Bias}
            Motivated by the above, we aim to separate semantic granularity for each level to address the low rationality issue.
            Unfortunately, it is \textbf{\emph{non-trivial}} since semantic granularity is unknown and varies in each domain.
            Some studies borrow external knowledge graphs~\cite{wang2022knowledge,duan2023bayesian},
            but such auxiliary information cannot fit various domains and mostly are unavailable.
            To overcome this challenge, we propose a new Context-aware Disentangled Decoder (CDD).
            \Cref{fig_decoder_disentangled} illustrates CDD.

            To separate semantic granularity,
            we propose to introduce a contextual topical bias to the decoding of each level.
            We denote this bias as a learnable variable $\mbf{b}^{(\ell)} \! \in \! \mathbb{R}^{V} $ for level $\ell$.
            We expect it to contain the topical semantics from the contextual levels of level $\ell$ in a hierarchy,
            so that level $\ell$ turns to cover other different semantics.
            Let $\mbf{p}^{(\ell)}$ denote such topical semantics of level $\ell$,
            and we model it as
            \begin{align}
                \mbf{p}^{(\ell)} = \!\!\!\!\!\!\! \sum_{\ell' \! \in \! \{ \ell \!-\! 1, \ell \!+\! 1 \}}  \sum_{k=1}^{K^{(\ell')}} \mathrm{topK}( \mbf{\beta}^{(\ell')}_{k}, N_{\mscript{top}} ) . \label{eq_topical_semantic}
            \end{align}
            Here $\mathrm{topK}(\cdot, \cdot)$ returns a vector that retains the top $N_{\mscript{top}}$ elements of $\mbf{\beta}^{(\ell')}_{k}$ and fills the rest with $0$.
            As such, $\mbf{p}^{(\ell)}$ represents the contextual topical semantics as it includes the top related words of all topics at level $\ell\!-\!1$ and $\ell\!+\!1$
            (only involves level $\ell\!+\!1$ ($\ell\!-\!1$) if level $\ell$ is the top-level (lowest-level)).
            Then we assign these contextual topical semantics
            to the
            bias $\mbf{b}^{(\ell)}$:
            \begin{align}
                b^{(\ell)}_{i} = p^{(\ell)}_{i} \quad \text{where} \quad p^{(\ell)}_{i} \neq 0. \label{eq_bias}
            \end{align}
            So $\mbf{b}^{(\ell)}$ contains the topical semantics from the contextual levels
            and also allows flexible bias learning on the semantics \emph{not} covered by these levels.
            See an example in \Cref{fig_decoder_disentangled}.

        \paragraph{Disentangled Decoding with Contextual Topical Bias}
            Instead of entangled decoding as early,
            we disentangle the decoding for each level with contextual topical biases.
            To be specific,
            we decode the document $\mbf{x}$ with topics at level $\ell$ by sampling word $x$ from a Multinomial distribution:
            \begin{align}
                x  \sim  \mathrm{Multi}( \mathrm{softmax}( \mbf{\beta}^{(\ell)} \mbf{\theta}^{(\ell)} + \lambda_{\mscript{b}} \mbf{b}^{(\ell)} )) \label{eq_reconstruction}
            \end{align}
            Here $\mbf{\beta}^{(\ell)} \mbf{\theta}^{(\ell)}$
            is the unnormalized generation probabilities following \citet{Srivastava2017}.
            Recall that $\mbf{\beta}^{(\ell)}$ is the topic-word distribution matrix,
            and $\mbf{\theta}^{(\ell)}$ is the doc-topic distribution of $\mbf{x}$ at level $\ell$.
            The decoding incorporates the contextual topical bias $\mbf{b}^{(\ell)}$ with a weight hyperparameter $\lambda_{\mscript{b}}$,
            \ie it knows the topical semantics of contextual levels.
            Thus the decoding turns to assign $\mbf{\beta}^{(\ell)}$, topics at level $\ell$, with semantics different from contextual levels. 
            This explicitly separates different semantic granularity and properly distributes them to topics at different levels.
            As a result, we can effectively improve the rationality of hierarchies
            See evidence in ablation studies.

    \subsection{Transport Plan and Context-aware Hierarchical Topic Model}
        Finally we formulate the objective for our Transport Plan and Context-aware Hierarchical Topic Model (TraCo).
        \paragraph{Objective for Topic Modeling}
            Following the ELBO of VAE \cite{Kingma2014a},
            we write the topic modeling objective with \Cref{eq_reconstruction} as
            \begin{align}
                \mathcal{L}_{\mscript{TM}}(\mbf{x}) =
                & \frac{1}{L} \sum_{\ell=1}^{L} \!\! - \mbf{x}^{\top} \! \log \bigl( \mathrm{softmax}(\mbf{\beta}^{(\ell)} \mbf{\theta}^{(\ell)} \!+\! \lambda_{\mscript{b}} \mbf{b}^{(\ell)}) \bigr) \notag
                \\
                & + \mathrm{KL} \Bigl[ q(\mbf{r} | \mbf{x}) \| p(\mbf{r}) ) \Bigr]
                \label{eq_TM}
            \end{align}
            The first term measures the average reconstruction error over all levels;
            the second term is the KL divergence between the prior and variational distributions.

        \paragraph{Objective for TraCo}
            Based on the above,
            we write the overall objective for TraCo by combining \Cref{eq_TPD,eq_TM}:
            \begin{align}
                \min_{\Theta, \mbf{W}, \{ \mbf{T}^{(\ell)} \}_{\ell=1}^{L} } \!\!\!\!\! \lambda_{\mscript{TPD}} \frac{1}{L\!-\!1} \! \sum_{\ell=1}^{L-1} \! \mathcal{L}^{(\ell)}_{\mscript{TPD}}
                \!+\!
                \frac{1}{N} \sum_{i=1}^{N} \! \mathcal{L}_{\mscript{TM}}(\mbf{x}^{(i)})
                \label{eq_overall}
            \end{align}
            where $\lambda_{\mscript{TPD}}$ is a weight hyperparameter.
            Here $\mathcal{L}^{(\ell)}_{\mscript{TPD}}$
            regularizes topic hierarchy building with sparse and balanced dependencies;
            $\mathcal{L}_{\mscript{TM}}$ assigns topics at each level with different semantic granularity and infers doc-topic distributions.

\begin{table*}[ht]
    \centering
    \setlength{\tabcolsep}{1.5mm}
    \renewcommand{\arraystretch}{1.1}
    \resizebox{\linewidth}{!}{
    \begin{tabular}{lrrrrrrrrrrrrrrrrrrrrrrrr}
        \toprule
        \multirow{2}[4]{*}{Model} & \multicolumn{4}{c}{NeurIPS}   &       & \multicolumn{4}{c}{ACL}       &       & \multicolumn{4}{c}{NYT}       &       & \multicolumn{4}{c}{Wikitext-103} &       & \multicolumn{4}{c}{20NG} \\
        \cmidrule{2-5}\cmidrule{7-10}\cmidrule{12-15}\cmidrule{17-20}\cmidrule{22-25}      & \multicolumn{1}{c}{PCC} & \multicolumn{1}{c}{PCD} & \multicolumn{1}{c}{SD} & \multicolumn{1}{c}{PnCD} &       & \multicolumn{1}{c}{PCC} & \multicolumn{1}{c}{PCD} & \multicolumn{1}{c}{SD} & \multicolumn{1}{c}{PnCD} &       & \multicolumn{1}{c}{PCC} & \multicolumn{1}{c}{PCD} & \multicolumn{1}{c}{SD} & \multicolumn{1}{c}{PnCD} &       & \multicolumn{1}{c}{PCC} & \multicolumn{1}{c}{PCD} & \multicolumn{1}{c}{SD} & \multicolumn{1}{c}{PnCD} &       & \multicolumn{1}{c}{PCC} & \multicolumn{1}{c}{PCD} & \multicolumn{1}{c}{SD} & \multicolumn{1}{c}{PnCD} \\
        \midrule
        nTSNTM & -0.348\sig & 0.603\sig & 0.195\sig & 0.566\sig &       & -0.214\sig & 0.674\sig & 0.268\sig & 0.653\sig &       & -0.450\sig & 0.501\sig & 0.193\sig & 0.479\sig &       & -0.026\sig & 0.816\sig & 0.500\sig & 0.777\sig &       & -0.089\sig & 0.745\sig & 0.323\sig & 0.765\sig \\
        HNTM  & -0.214\sig & 0.719\sig & 0.410\sig & 0.775\sig &       & -0.095\sig & 0.867\sig & 0.568\sig & 0.887\sig &       & -0.137\sig & 0.757\sig & 0.380\sig & 0.723\sig &       & -0.190\sig & 0.903\sig & 0.637\sig & 0.941\sig &       & -0.332\sig & 0.832\sig & 0.425\sig & 0.796\sig \\
        NGHTM & 0.014\sig & 0.905\sig & 0.635\sig & 0.954\sig &       & 0.055\sig & 0.902\sig & 0.633\sig & 0.947\sig &       & -0.026\sig & 0.816\sig & 0.351\sig & 0.887\sig &       & 0.054\sig & 0.933\sig & 0.548\sig & 0.956\sig &       & -0.011\sig & 0.831\sig & 0.446\sig & 0.863\sig \\
        SawETM & -0.093\sig & 0.785\sig & 0.816\sig & 0.986\sig &       & -0.095\sig & 0.772\sig & 0.782\sig & 0.977\sig &       & -0.234\sig & 0.641\sig & 0.680\sig & 0.970\sig &       & -0.190\sig & 0.709\sig & 0.683\sig & 0.931\sig &       & -0.332\sig & 0.563\sig & 0.543\sig & 0.945\sig \\
        DCETM & -0.361\sig & 0.605\sig & 0.485\sig & 0.858\sig &       & -0.353\sig & 0.584\sig & 0.387\sig & 0.804\sig &       & -0.041\sig & 0.802\sig & 0.756\sig & 0.978\sig &       & -0.522\sig & 0.471\sig & 0.344\sig & 0.506\sig &       & -0.085\sig & 0.742\sig & 0.644\sig & 0.900\sig \\
        ProGBN & -0.119\sig & 0.746\sig & 0.576\sig & 0.976\sig &       & -0.058\sig & 0.781\sig & 0.611\sig & 0.976\sig &       & -0.049\sig & 0.753\sig & 0.614\sig & 0.983\sig &       & 0.068\sig & 0.885\sig & 0.707\sig & 0.983\sig &       & -0.009\sig & 0.780\sig & 0.626\sig & 0.981\sig \\
        HyperMiner & -0.084\sig & 0.771\sig & 0.808\sig & 0.991\sig &       & -0.063\sig & 0.757\sig & 0.824\sig & 0.990\sig &       & -0.229\sig & 0.638\sig & 0.713\sig & 0.984\sig &       & -0.207\sig & 0.703\sig & 0.685\sig & 0.949\sig &       & -0.256\sig & 0.604\sig & 0.584\sig & 0.959\sig \\
        \midrule
        \textbf{TraCo} & \textbf{0.077} & \textbf{0.958} & \textbf{0.972} & \textbf{0.999} &       & \textbf{0.081} & \textbf{0.932} & \textbf{0.967} & \textbf{0.999} &       & \textbf{-0.021} & \textbf{0.946} & \textbf{0.946} & \textbf{0.998} &       & \textbf{0.167} & \textbf{0.947} & \textbf{0.960} & \textbf{0.999} &       & \textbf{0.037} & \textbf{0.895} & \textbf{0.894} & \textbf{0.997} \\
        \bottomrule
    \end{tabular}%
    }
    \caption{
        Topic hierarchy quality results.
        PCC and PCD refer to the coherence and diversity between parent and child topics respectively;
        PnCD is the diversity between parent and non-child topics;
        SD is the diversity between sibling topics.
        The best are in bold.
        The superscript $\ddagger$ means the gain of TraCo is statistically significant at 0.05 level.
    }
    \label{tab_topic_hierarchy_quality}%
\end{table*}%

\begin{table*}[!ht]
    \centering
    \setlength{\tabcolsep}{1.5mm}
    \renewcommand{\arraystretch}{1.1}
    \resizebox{\linewidth}{!}{
    \begin{tabular}{lrrrrrrrrrrrrrrrrrrrrrrrr}
        \toprule
        \multirow{2}[4]{*}{Model} & \multicolumn{4}{c}{NeurIPS}   &       & \multicolumn{4}{c}{ACL}       &       & \multicolumn{4}{c}{NYT}       &       & \multicolumn{4}{c}{Wikitext-103} &       & \multicolumn{4}{c}{20NG} \\
        \cmidrule{2-5}\cmidrule{7-10}\cmidrule{12-15}\cmidrule{17-20}\cmidrule{22-25}      & \multicolumn{1}{c}{PCC} & \multicolumn{1}{c}{PCD} & \multicolumn{1}{c}{SD} & \multicolumn{1}{c}{PnCD} &       & \multicolumn{1}{c}{PCC} & \multicolumn{1}{c}{PCD} & \multicolumn{1}{c}{SD} & \multicolumn{1}{c}{PnCD} &       & \multicolumn{1}{c}{PCC} & \multicolumn{1}{c}{PCD} & \multicolumn{1}{c}{SD} & \multicolumn{1}{c}{PnCD} &       & \multicolumn{1}{c}{PCC} & \multicolumn{1}{c}{PCD} & \multicolumn{1}{c}{SD} & \multicolumn{1}{c}{PnCD} &       & \multicolumn{1}{c}{PCC} & \multicolumn{1}{c}{PCD} & \multicolumn{1}{c}{SD} & \multicolumn{1}{c}{PnCD} \\
        \midrule
        w/o TPD & -0.033\sig & 0.920\sig & 0.710\sig & 0.991\sig &       & -0.023\sig & 0.898\sig & 0.606\sig & 0.968\sig &       & -0.162\sig & 0.930\sig & 0.821\sig & 0.993\sig &       & -0.286\sig & 0.665\sig & 0.452\sig & 0.790\sig &       & -0.084\sig & \textbf{0.900} & 0.723\sig & 0.985\sig \\
        w/o CDD & -0.083\sig & 0.772\sig & 0.907\sig & 0.996\sig &       & -0.034\sig & 0.795\sig & 0.905\sig & 0.996\sig &       & -0.140\sig & 0.731\sig & 0.828\sig & 0.991\sig &       & 0.030\sig & 0.802\sig & 0.918\sig & 0.997\sig &       & -0.145\sig & 0.719\sig & 0.780\sig & 0.990\sig \\
        \midrule
        \textbf{TraCo} & \textbf{0.077} & \textbf{0.958} & \textbf{0.972} & \textbf{0.999} &       & \textbf{0.081} & \textbf{0.932} & \textbf{0.967} & \textbf{0.999} &       & \textbf{-0.021} & \textbf{0.946} & \textbf{0.946} & \textbf{0.998} &       & \textbf{0.167} & \textbf{0.947} & \textbf{0.960} & \textbf{0.999} &       & \textbf{0.037} & 0.895 & \textbf{0.894} & \textbf{0.997} \\
        \bottomrule
    \end{tabular}%
    }
    \caption{
        Ablation study:
        without Transport Plan Dependency (w/o TDP);
        without Context-aware Disentangled Decoder (w/o CDD).
        The best are in bold.
        The superscript $\ddagger$ means the gain of TraCo is statistically significant at 0.05 level.
    }
    \label{tab_ablation}%
\end{table*}%

\section{Experiment}
    In this section we conduct experiments to show the effectiveness of our method.

    \subsection{Experiment Setup}
        \paragraph{Datasets}
            We experiment with the following benchmark datasets:
            \begin{inparaenum}[(i)]
                \item \textbf{NeurIPS}
                contains the publications at the NeurIPS conference from 1987 to 2017.
                \item \textbf{ACL}
                \cite{bird2008acl}
                is a paper collection from the ACL anthology from 1970 to 2015.
                \item \textbf{NYT}
                contains news articles of the New York Times with 12 categories.
                \item \textbf{Wikitext-103}
                \cite{merity2016pointer}
                includes Wikipedia articles.
                \item \textbf{20NG}
                \cite{Lang95}
                includes news articles with 20 labels.
            \end{inparaenum}

        \paragraph{Baseline Models}
            We consider the following state-of-the-art baseline models:
            \begin{inparaenum}[(i)]
                \item
                    \textbf{nTSNTM} \cite{chen2021tree} uses a stick-breaking process prior.
                \item
                    \textbf{HNTM} \cite{chen2021hierarchical} introduces manifold regularization on topic dependencies.
                \item
                    \textbf{SawETM} \cite{duan2021sawtooth} proposes a Sawtooth Connection to model topic dependencies.
                \item
                    \textbf{DCETM} \cite{li2022alleviating} uses skip-connections in document decoding
                    and a policy gradient training approach.
                \item
                    \textbf{HyperMiner}~\cite{xu2022hyperminer} projects topic and word embeddings into hyperbolic space.
                \item
                    \textbf{NGHTM} \cite{chen2023nonlinear} models dependencies via non-linear equations.
                 \item
                    \textbf{ProGBN} \cite{duan2023bayesian} crafts documents with more related words for the decoding of higher levels.
            \end{inparaenum}
            We report average results of 5 runs.
            See more implementation details in the Appendix.

    \subsection{Topic Quality}
        \paragraph{Evaluation Metrics}
            We adopt the below metrics following normal topic quality evaluation:
            \begin{inparaenum}[(i)]
                \item
                    \textbf{T}opic \textbf{C}oherence (\textbf{TC}) measures the coherence between top words of topics.
                    We evaluate with the widely-used metric $C_V$
                    ,
                    outperforming earlier ones \cite{Newman2010,roder2015exploring}.
                \item
                    \textbf{T}opic \textbf{D}iversity (\textbf{TD})
                    refers to differences between topics.
                    Following \citet{dieng2020topic}, we measure TD as
                    the uniqueness of top related words in topics.
            \end{inparaenum}

        \paragraph{Result Analysis}
            \Cref{tab_topic_quality} shows the average TC and TD scores over all levels.
            We see our TraCo consistently outperforms baselines concerning both TC and TD.
            Especially TraCo achieves significantly higher TD scores.
            For example, TraCo reaches a TD score of 0.824 on NeurIPS while the runner-up only has 0.632.
            These results demonstrate that our model can generate high-quality topics for different levels with better coherence and diversity.

\begin{figure*}[!t]
    \centering
    \includegraphics[width=0.9\linewidth]{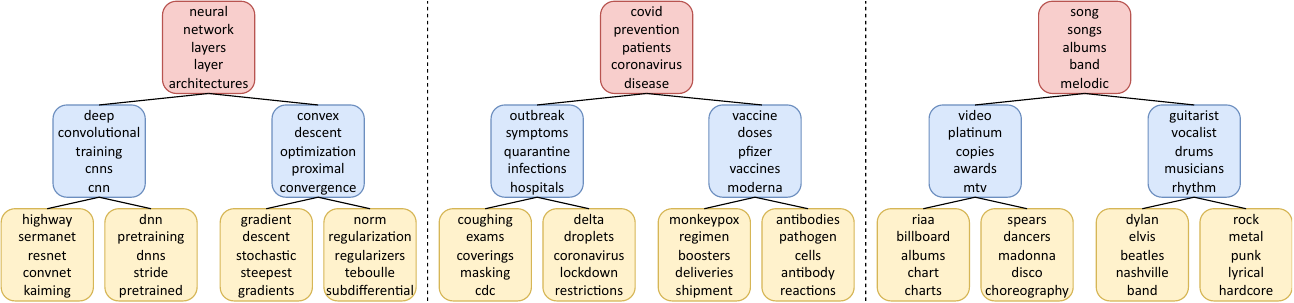}
    \caption{
        Case study: discovered topic hierarchies from different datasets.
        Each rectangle is the top related words of a topic.
    }
    \label{fig_case_study}
\end{figure*}

    \subsection{Topic Hierarchy Quality} \label{sec_topic_hierarchy_quality}
        \paragraph{Evaluation Metrics}
            We consider the following metrics to evaluate topic hierarchy:
            \begin{inparaenum}[(i)]
                \item
                    \textbf{P}arent and \textbf{C}hild Topic \textbf{C}oherence (\textbf{PCC}) indicates the coherence between parent and child topics.
                    We use CLNPMI \cite{chen2021tree} to measure it.
                    CLNPMI computes the NPMI \cite{lau2014machine} of every two
                    words from a parent topic and its child topic.
                \item
                    \textbf{P}arent and \textbf{C}hild Topic \textbf{D}iversity (\textbf{PCD}) measures the diversity between a parent topic and its child \cite{chen2021tree}.
                    PCC and PCD together verify if parent and child topics are relevant and cover different semantic granularity.
                    This evaluates the rationality of a topic hierarchy.
                \item
                    \textbf{P}arent and \textbf{n}on-\textbf{C}hild Topic \textbf{D}iversity (\textbf{PnCD}) measures the diversity between a parent topic and its non-child \cite{isonuma2020tree,chen2021hierarchical}.
                    It verifies whether a child topic only has a high affinity to its parent topic.
                \item
                    \textbf{S}ibling Topic \textbf{D}iversity (\textbf{SD}) measures the diversity between sibling topics.
                    Note that PCD cannot replace SD since a parent topic may have repeating children.
            \end{inparaenum}
            We follow the TD metric \cite{dieng2020topic} to compute the above PCD, PnCD, and SD.

        \paragraph{Result Analysis}
            \Cref{tab_topic_hierarchy_quality} reports the topic hierarchy quality results.
            We have the following observations:
            \begin{inparaenum}[(\bgroup\bfseries i\egroup)]
                \item
                    \textbf{Our model shows higher affinity.}
                    We see that our TraCo significantly surpasses all baselines concerning PCC and PnCD.
                    This signifies that parent topics more relate to their children and differ from non-children in the hierarchies of TraCo,
                    manifesting its enhanced affinity.
                \item
                    \textbf{Our model attains better rationality.}
                    Besides the best PCC, our TraCo reaches the best PCD compared to all baselines.
                    For example, TraCo has PCC of 0.077 and PCD of 0.958 on NeurIPS
                    while the runner-up has 0.014 and 0.905.
                    This evidences that parent and child topics contain not only related semantics but also different granularity,
                    which shows higher rationality of our method.
                \item
                    \textbf{Our model achieves higher diversity.}
                    \Cref{tab_topic_hierarchy_quality} shows our TraCo outperforms baselines in terms of SD.
                    For example,
                    NGHTM has a close PCC score on NYT, but TraCo reaches much higher SD (0.946 vs. 0.351).
                    This demonstrates our model produces more diverse sibling topics instead of repetitive ones.
            \end{inparaenum}

    \subsection{Ablation Study} \label{sec_ablation_study}
        We conduct ablation studies to show the necessity of our TPD and CDD methods.
        From \Cref{tab_ablation},
        we see that \textbf{TPD effectively mitigates the low affinity and diversity issues.}
        PCC and SD scores degrade largely if without TPD (w/o TPD).
        For example, PCC decreases from 0.167 to -0.286 and SD from 0.960 to 0.452 on Wikitext-103.
        This implies less related parent and child topics and repetitive siblings.
        These results verify that our TPD facilitates the affinity and diversity of topic hierarchies.
        Besides,
        we notice that \textbf{CDD can alleviate the low rationality issue.}
        PCC and PCD decline significantly if without CDD (w/o CDD), like from 0.081 to -0.034 and from 0.932 to 0.795 on ACL,
        indicating less distinguishable parent and child topics.
        This demonstrates that our CDD improves the rationality of topic hierarchies.

    \subsection{Text Classification and Clustering}
        Apart from the above comparisons,
        we evaluate inferred doc-topic distributions through
        downstream tasks: text classification and clustering.
        Specifically,
        we train SVM classifiers with learned doc-topic distributions as features and predict document labels,
        evaluated by Accuracy (Acc) and F1.
        For clustering,
        we use the most significant topics in doc-topic distributions as clustering assignments, evaluated by Purity and NMI following \citet{zhao2020neural}.
        We take the average
        classification and clustering
        results over all hierarchy levels
        on the NYT and 20NG datasets.

\begin{figure}[!t]
    \centering
    \begin{subfigure}[c]{0.5\linewidth}
        \centering
        \includegraphics[width=\linewidth]{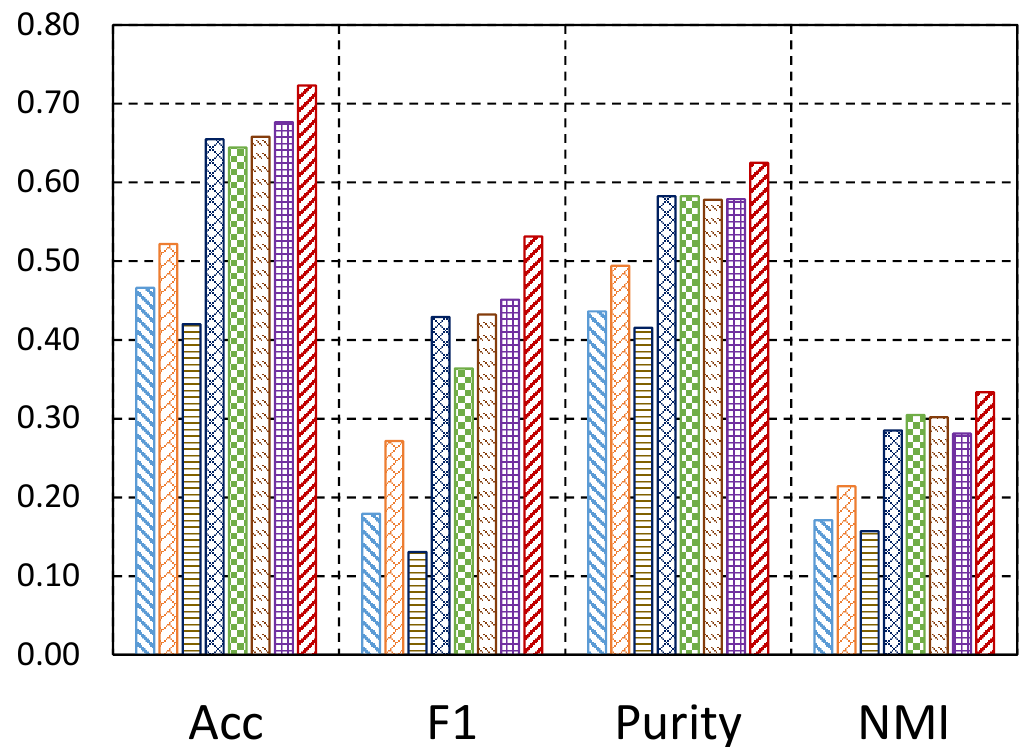}
        \caption{NYT}
        \label{fig_classification_NYT}
    \end{subfigure}%
    \begin{subfigure}[c]{0.5\linewidth}
        \centering
        \includegraphics[width=\linewidth]{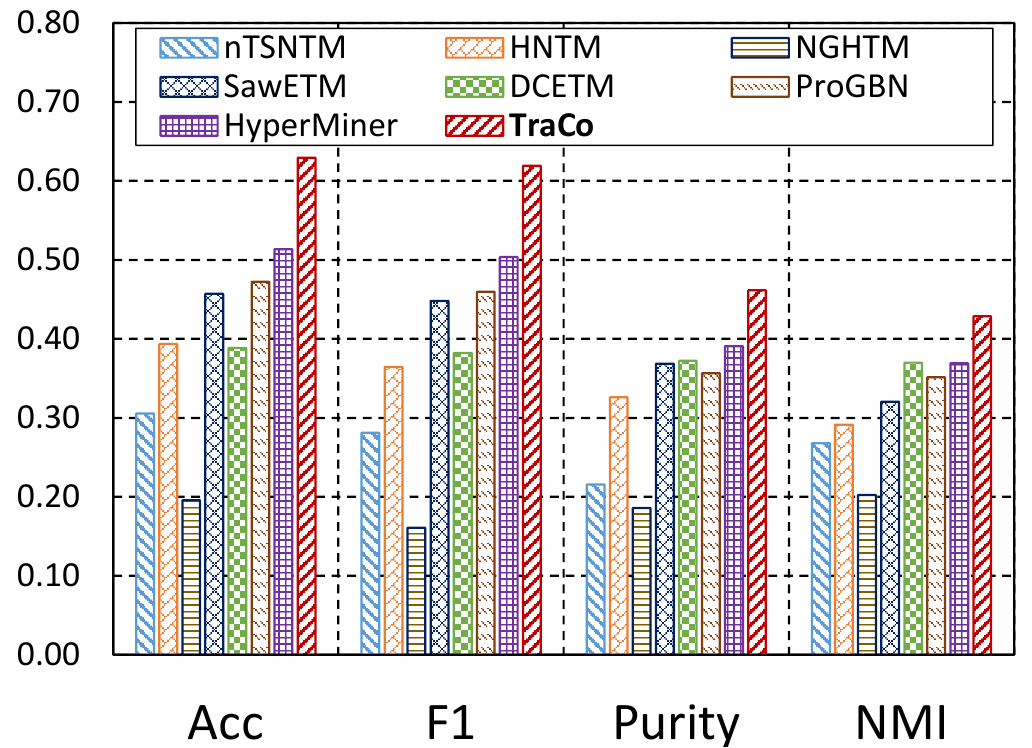}
        \caption{20NG}
        \label{fig_classification_20NG}
    \end{subfigure}%
    \caption{
        Text classification (Acc and F1) and clustering results (Purity and NMI).
        The gains of our TraCo are all statistically significant at 0.05 level.
    }
    \label{fig_classification}
\end{figure}

        \Cref{fig_classification} shows our TraCo consistently outperforms baseline methods in terms of both text classification and clustering.
        These demonstrate that our model can infer higher-quality doc-topic distributions for different hierarchy levels, which can benefit downstream applications.
        As we infer higher-level doc-topic distributions via dependencies (\Cref{eq_theta}),
        these manifest that the learned dependencies of our model are accurate as well.

    \subsection{Case Study: Discovered Topic Hierarchy} \label{sec_case_study}
        We conduct case studies to illustrate our model discovers affinitive, rational, and diverse topic hierarchies.
        \Cref{fig_case_study} exemplifies discovered topic hierarchies by our model from NeurIPS, NYT, and Wikitext-103.
        Specifically,
        the left part of \Cref{fig_case_study} shows a general parent topic relates to ``neural network'',
        associated with affinitive offspring topics like ``cnn'', ``resnet'', ``optimization''.
        The middle part
        of \Cref{fig_case_study}
        illustrates a general parent topic related to ``covid''
        and specific offspring topics about ``symptoms'', ``lockdown'', and ``vaccine'' with children like ``booster'' and ``antibodies''.
        Moreover, the right part
        of \Cref{fig_case_study}
        presents a general parent topic focusing on ``songs'',
        and specific offspring on ``albums'', musicians like ``dylan'' and ``beatles'', and music genres like ``rock'', ``metal'', and ``punk''.

\section{Conclusion}
    In this paper we propose TraCo for hierarchical topic modeling.
    Our TraCo uses a transport plan dependency method to address the low affinity and diversity issues,
    and leverages a context-aware disentangled decoder to mitigate the low rationality issue.
    Experiments demonstrate that TraCo can consistently outperform baselines,
    producing higher-quality topic hierarchies with significantly improved affinity, diversity, and rationality.
    Especially TraCo shows better performance on downstream tasks with more accurate topic distributions of documents.

\section*{Acknowledgements}
    We thank all anonymous reviewers for their helpful comments.
    This research/project is supported by the National Research Foundation, Singapore under its AI Singapore Programme, AISG Award No: AISG2-TC-2022-005.

\bibliography{lib}

\clearpage

\appendix

\begin{algorithm}
    \caption{Training algorithm for TraCo.}
    \label{training_algorithm}
    \textbf{Input:} document collection $\{ \mbf{x}^{(1)}, \dots, \mbf{x}^{(N)} \}$; \\
    \textbf{Output:} model parameters $\Theta$, $\mbf{W}$, $\{\mbf{T}^{(\ell)}\}_{\ell=1}^{L}$; \\
    \begin{algorithmic}[1]
        \FOR{ 1 \textbf{to} $n_{\mscript{epoch}}$ }
            \FOR{ $\ell=1$ \textbf{to} $L-1$ }
                \STATE \COMMENT{Sinkhorn's algorithm};
                \STATE $ C^{((\ell)}_{kk'} = \| \mbf{t}^{(\ell+1)}_{k} - \mbf{t}^{(\ell)}_{k'} \|^{2} $;
                \STATE $ \mbf{M} = \exp( -\mbf{C}^{(\ell)} / \varepsilon ) $;
                \STATE $ \mbf{b} \leftarrow \mbf{\mathds{1}}_{K^{(\ell)}} $;
                \WHILE{not converged and not reach max iterations}
                    \STATE $ \mbf{a} \leftarrow \frac{1}{K^{(\ell+1)}} \frac{\mbf{\mathds{1}}_{K^{(\ell+1)}}}{\mbf{M}} \mbf{b} $
                    , $ \mbf{b} \leftarrow \frac{\mbf{s}^{(\ell)}}{\mbf{M}^{\top} \mbf{a} }$;
                \ENDWHILE
                \STATE Compute $\mbf{\varphi}^{(\ell)}\leftarrow \diag(\mbf{a}) \mbf{M} \diag(\mbf{b})$;
            \ENDFOR
            \STATE Compute \Cref{eq_overall};
            \STATE Update $\Theta$, $\mbf{W}$, $\{\mbf{T}^{(\ell)}\}_{\ell=1}^{L}$ with a gradient step;
        \ENDFOR
    \end{algorithmic}
\end{algorithm}

\section{Training Algorithm for TraCo} \label{sec_app_training_algorithm}
    \Cref{training_algorithm} shows the training algorithm for our TraCo.
    We use Sinkhorn's algorithm \cite{sinkhorn1964relationship,cuturi2013lightspeed} to obtain $\mbf{\varphi}^{(\ell)}$, the approximated optimal transport plan solution of \Cref{eq_OT_hierarchical_topic} as the dependencies between topics at level $\ell$ and $\ell \!+\! 1$.

\section{Dataset Pre-processing} \label{sec_app_datasets}
    To pre-process datasets, we follow the steps in \citet{Card2018a,wu2023effective}:
    \begin{inparaenum}[(1)]
        \item
            tokenize documents and convert them to lowercase;
        \item
            remove punctuation;
        \item
            remove tokens that include numbers;
        \item
            remove tokens less than 3 characters;
        \item
            remove stop words.
    \end{inparaenum}

\section{Implementation Details} \label{sec_app_implementation}
    Following \citet{chen2021hierarchical}, we set a 3-level topic hierarchy for experiments, each with 10, 50, and 200 topics.
    For Sinkhorn's algorithm, we set the maximum number of iterations as 1,000, the stop tolerance 0.005, and $\varepsilon$ 0.05 following \citet{cuturi2013lightspeed}.
    We set $\tau$ in \Cref{eq_beta} as 0.1,
    $N_{\mscript{top}}$ in \Cref{eq_topical_semantic} as 20,
    $\lambda_{\mscript{b}}$ in \Cref{eq_TM} as 5.0,
    and
    $\lambda_{\mscript{TPD}}$ in \Cref{eq_overall} as 20.0.
    Following \cite{wu2023effective},
    our encoder network is a MLP that has two linear layers with a softplus activation function, concatenated with two single layers each for the mean and covariance matrix.
    We use Adam \cite{Kingma2014} to optimize model parameters with a learning rate of 0.002 and 200 epochs.

\end{document}